\documentclass{article}

\usepackage{tikz}
\usepackage{pgfplots}
\usepackage{gensymb}
\usepackage{amsfonts}

\usepackage[
  backend=bibtex,
  bibstyle=ieee,
  citestyle=numeric-comp,
  natbib=true,
  mincitenames=1,
  maxcitenames=1,
]{biblatex}
\usepackage{spconf,amsmath,epsfig}

\usepackage{subcaption}
\usepackage{siunitx}
\DeclareSIUnit\px{px}

\usepackage{hyperref}

\usepackage[%
  xindy,
  acronym,
]{glossaries}
\glsdisablehyper
\defglsdisplayfirst[acronym]{\emph{#1#4}}

\usepackage{tumabbrev}

\title{Map-Repair: Deep cadastre Maps Alignment and Temporal Inconsistencies Fix in Satellite Images}
\name{Stefano Zorzi\textsuperscript{1}, Ksenia Bittner\textsuperscript{2},
        Friedrich Fraundorfer\textsuperscript{1}}
\address{\textsuperscript{1 }Institute of Computer Graphics and Vision, Graz University of Technology, Graz, Austria \\
(stefano.zorzi, fraundorfer)@icg.tugraz.at \\
\textsuperscript{2 }Remote Sensing Technology Institute, German Aerospace Center (DLR), Wessling, Germany \\  
ksenia.bittner@dlr.de }
\usepackage{fancyhdr}
\begin{document}
\lhead{\centering This manuscript was accepted to be presented at the IEEE International Geoscience and Remote Sensing Symposium 2020.}
\thispagestyle{fancy}

\maketitle

\begin{abstract}

In the fast developing countries it is hard to trace new buildings construction or old structures destruction and, as a result, to keep the up-to-date cadastre maps.
Moreover, due to the complexity of urban regions or inconsistency of data used for cadastre maps extraction, the errors in form of misalignment is a common problem. 
In this work, we propose an end-to-end deep learning approach which is able to solve inconsistencies between the input intensity image and the available building footprints by correcting label noises and, at the same time, misalignments if needed. 
%by adding or removing not considering building objects, and correcting at the same time the misalignment if needed.  
The obtained results demonstrate the robustness of the proposed method to even severely misaligned examples that makes it potentially suitable for real applications, like \emph{OpenStreetMap} correction.
\end{abstract}

\section{Introduction}

\label{sec:intro}
Semantic segmentation is still a challenging problem in Remote Sensing.
Automatic detection and extraction of precise object outlines, such as human constructions, is in the interest of many cartographic and engineering applications.
The most effective way to deal with this problem is the use of \emph{Convolutional Neural Networks} trained in a supervised manner.
Accurate ground truth annotations allows to achieve great detection and segmentation accuracies, however, these good annotations are hard to come by because they might be misaligned due to multiple causes e.g. human errors or imprecise digital terrain model.
Furthermore, the maps may not be temporally synchronized with the satellite images failing to take into account variations in the constructions, i.e. new buildings may have been built or destroyed.

Several related works tackle this problem with different approaches.
Good alignment performance are achieved in \cite{girard2018aligning} by training a CNN to predict a displacement field between a map and an image.
The same authors proposed in \cite{girard2019noisy} a multi-rounds training scheme which ameliorates ground truth annotations at each round to fine-tune the model.
More recently, a method that performs a sequential annotation adjustment using a combination of consistency and self-supervised losses has been published \cite{chen2019autocorrect}.

In this paper we propose an end-to-end self-supervised deep learning method for the generation of aligned and temporally coherent cadastre annotation in satellite and airborne imagery.
The aim of the method is to align in one single shot all the object instances present in the intensity image and, at the same time, detect obsolete footprints and segment constructions that lack annotations.

\begin{figure}[]
\centering
\includegraphics[width=\linewidth]{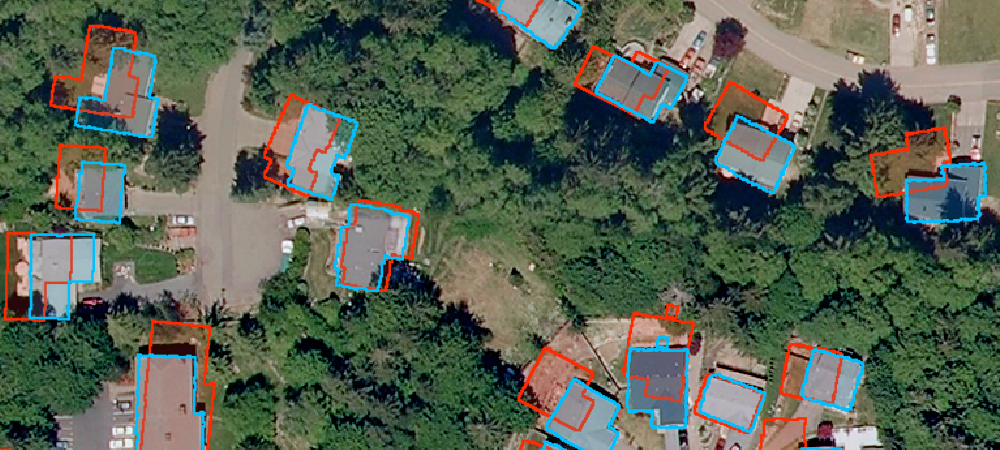}
\caption{\emph{MapRepair} result. Misaligned annotations in red, corrected map in cyan. }
\label{fig:result_kitsap36}
\end{figure}

\section{Methodology}
\label{sec:method}

%\begin{figure*}[thbp]
%\centering
%\includegraphics[width=1\linewidth]{imgs/map-repair.png}
%\put (-230,68) {\scriptsize{prediction from CNN}}
%\put (-150,68) {\scriptsize{ordering and filtering}}
%\put (-71,68) {\scriptsize{final polygon}}
%\caption{blablabla}
%\label{fig:workflow}
%\end{figure*}

\begin{figure*}
\centering
\tabskip=0pt
\valign{#\cr
  \hbox{%
    \begin{subfigure}[b]{.8\textwidth}
    \centering
    \includegraphics[width=\textwidth]{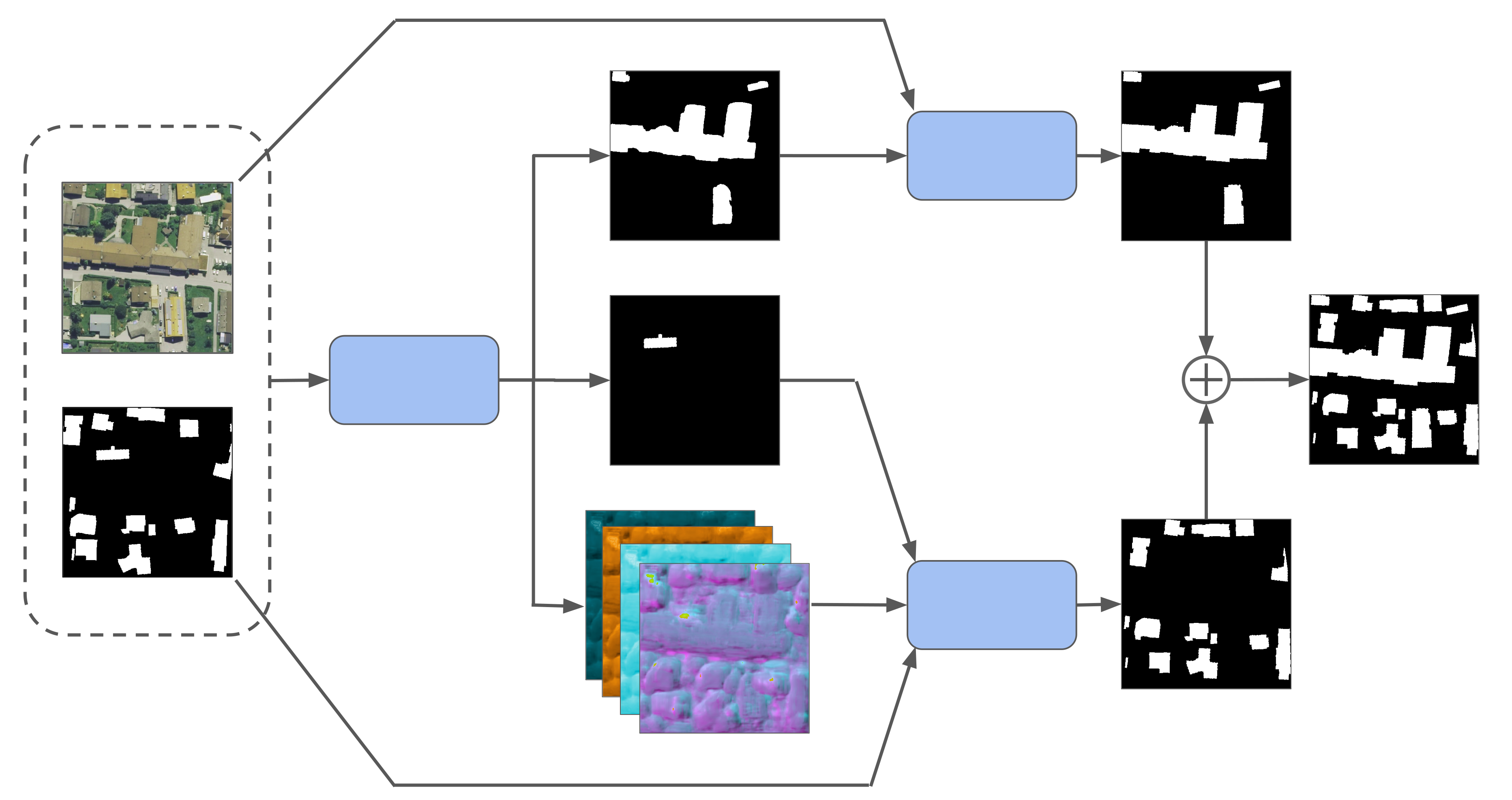}
    \end{subfigure}%
  }\cr
  \noalign{\hfill}
  \hbox{%
    \begin{subfigure}{.18\textwidth}
    \centering
    \includegraphics[width=\textwidth]{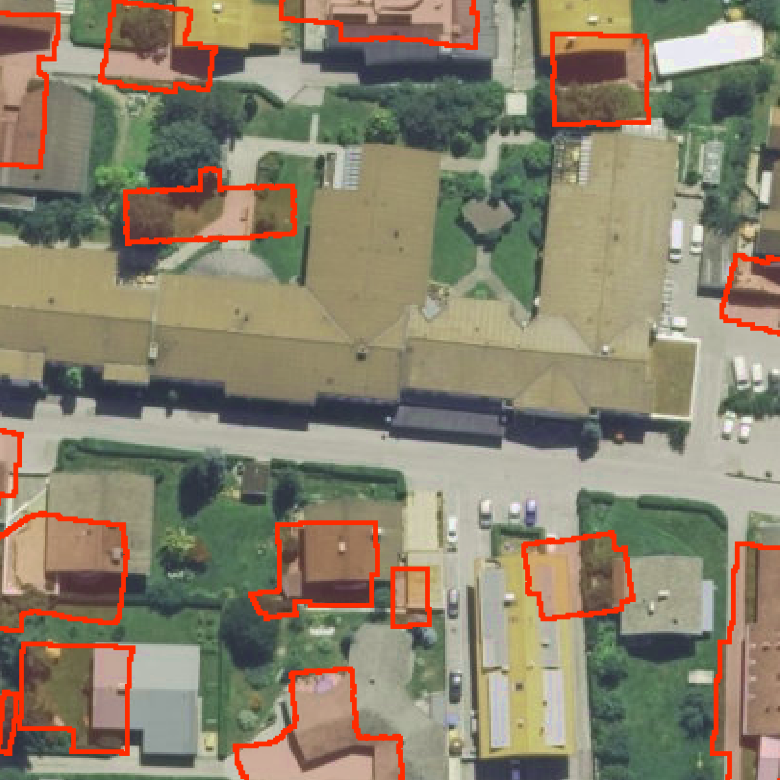}
    \caption{Noisy input}
    \end{subfigure}%
  }\vfill
  \hbox{%
    \begin{subfigure}{.18\textwidth}
    \centering
    \includegraphics[width=\textwidth]{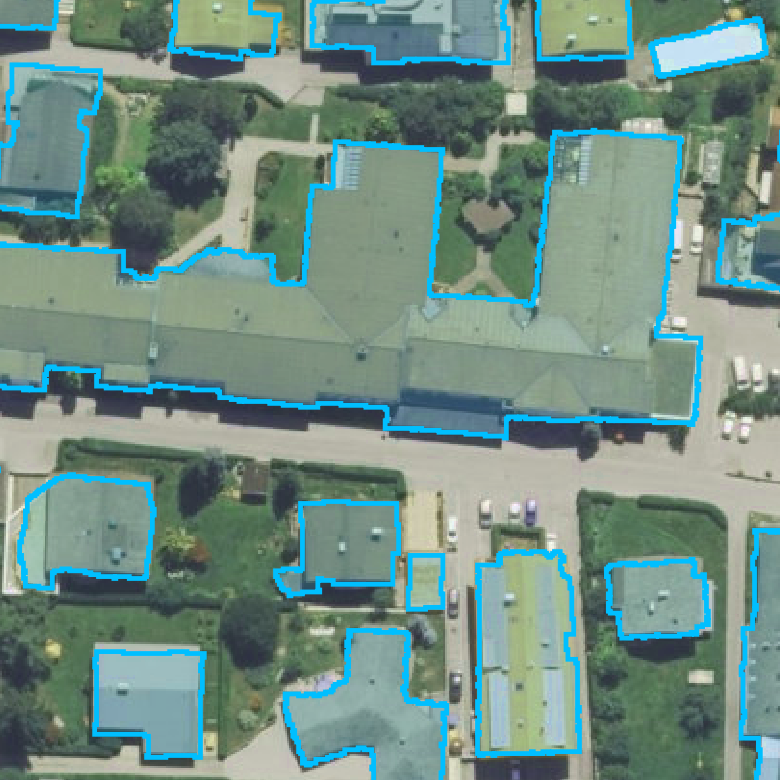}
    \caption{Final prediction}
    \end{subfigure}%
  }\cr
}
\put (-481,55) {\scriptsize{Noisy map}}
\put (-488,115) {\scriptsize{Intensity image}}
\put (-410,113) {\scriptsize{Generator}}
\put (-253,53) {\scriptsize{Warping}}
\put (-261,174) {\scriptsize{Regularization}}
\put (-331,145) {\scriptsize{Missing}}
\put (-335,85) {\scriptsize{To remove}}
\put (-332,14) {\scriptsize{Similarity map}}
\put (-200,25) {\scriptsize{Aligned map}}
\put (-145,85) {\scriptsize{Final map}}
\put (-208,202) {\scriptsize{Regularized map}}
\caption{Workflow of the proposed \emph{MapRepair} method. The intensity image and the noisy annotations are given to a network that generates a transformation map and segments missing and obsolete footprints. The aligned cadastre layer is produced by a warping function while the segmentation is improved by a regularization network. The final refined annotations are obtained merging the results from the regularization branch and the warping branch. Noisy and refined annotations overlaid to the RGB image are shown on the right side.}
\label{fig:workflow}
\end{figure*}
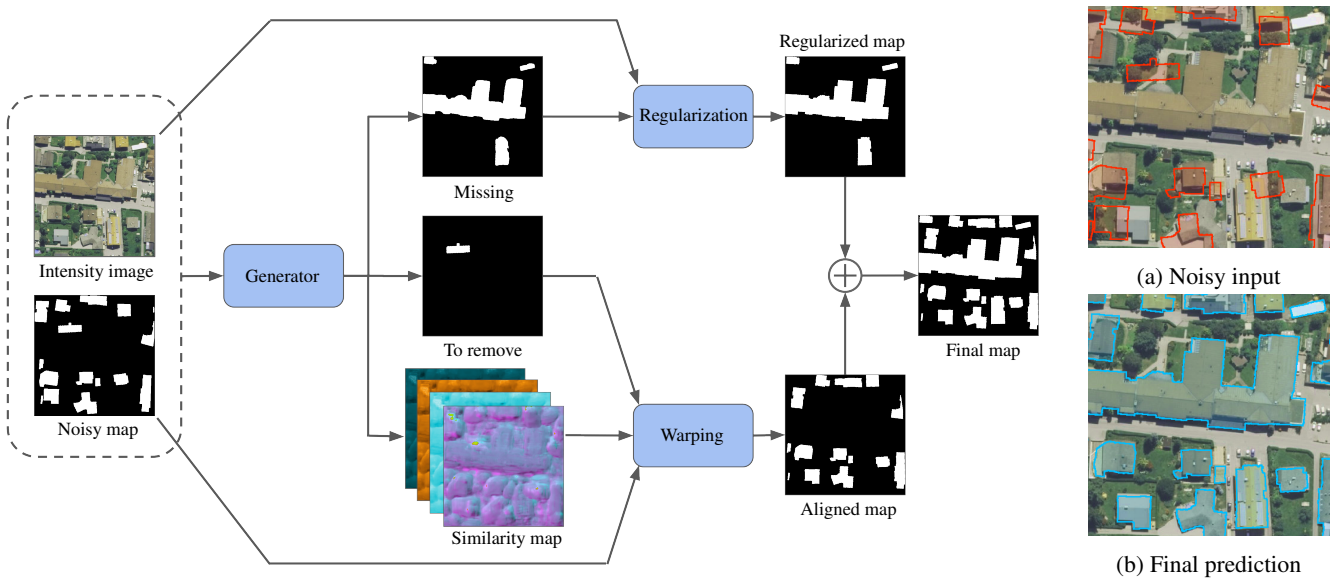

Our goal is to train a deep neural network which can not only generate an aligned cadastre map, but can also remove obsolete footprints and detect new buildings.
The overall network model is shown in Figure~\ref{fig:workflow}, and it is composed of two different branches.
The first branch estimates and performs a projection for every building instance in order to produce a map perfectly registered with the intensity image. 
During this process, obsolete footprints are discarded.
If a building does not have a footprint in the map, the second branch segments and regularizes the construction providing an accurate and visually pleasing building boundary.
The results from two paths are then merged to produce the final corrected map.

\subsection{Similarity map estimation and instance warping}
In order to better align individual object instances to the image content, a generator network $G$ is exploited to predict a similarity transformation map $T \in \mathbb{R}^{4 \times H \times W}$ where the channels store translation (along $x$ and $y$ axis), rotation and scale values for each pixel location.
The model receives as input the intensity image $I \in  \mathbb{R}^{3 \times H \times W}$ and a binary mask $y = \{0,1\}^{H \times W}$ which represents the noisy or misaligned annotations.

\begin{equation}
    T = G(I, y)
\end{equation}

A similarity transformation is then computed independently for every building averaging the values of the tensor $T$ under the area described by the object instance. 
The transformation for the $i$-th instance can be written as:

\begin{equation}
    t_i = \frac{1}{N_i} \sum_{p \in \omega_i} T_p 
\end{equation}
where $N_i$ and $\omega_i$ are the number of points and the set of points of the instance $i$-th, respectively.
The four values of $t_i$ define a $\mathbb{R}^2 \xrightarrow{} \mathbb{R}^2$ similarity transformation. 

The refined annotation for the $i$-th object instance $\hat{y}_i$ is expressed as the transformed version $\hat{y}_i=\text{warp}(y_i, t_i)$ of the noisy instance annotation $y_i$ by the predicted transformation $t_i$.

The predicted aligned annotations $\hat{y}$ for the binary image $y$ is then calculated as the combination of the single instance transformations and can be expressed as:

\begin{equation}
    \hat{y} = \sum_{i=1}^M \text{warp}(y_i, t_i)
\end{equation}
where $y_i$ represents the $i$-th instance of the noisy binary mask and $M$ is the number of object instances in the sample image. 

The loss function used to train the model to generate the similarity transformation map is a combination of the mean squared error and the mean absolute error between the predicted binary annotations $\hat{y}$ and the ground truth annotations.% $y_{gt}$:

\subsection{Segmentations and regularization}
Maps may not be temporally synchronized with the satellite or airborne data, failing to take into account variations in human constructions, i.e. removed or new buildings.

In order to solve this problem, the model $G$ also predicts two segmentation masks: the first represents footprints of buildings that lack of annotation in $y$, while the second shows the annotations that must be removed because obsolete.

The missing footprints predicted by $G$ have round corners and an irregular shape due to the lack of geometric constraints during the prediction.
In order to ameliorate the segmentation result we post-process the result with the regularization model proposed in \cite{8900337}.
This network for footprint refinement is capable of generating regular and visually pleasing building boundaries without losing segmentation accuracy.

The segmentation of the obsolete annotations is instead used by the warping function to filter out-of-date or wrong instances.

During training the ground truth of both the missing instances and the obsolete instances is known and binary cross-entropy losses are computed for these two segmentation branches.

The generator network $G$ is used both for the alignment task and for the detection task, therefore it is trained using a linear combination of the alignment losses and the segmentation losses.

\subsection{Network models}
The convolutional neural network used as generator $G$ is a recurrent residual U-Net~\cite{alom2018recurrent} modified to produce three outputs: two segmentation masks and the similarity transformation map.
The network we adopted is a simple but yet precise segmentation model which guarantees high building segmentation accuracy.
The input image has 4 channels, since it is the concatenation of the intensity image $I$ and the noisy annotation mask $y$. 
The outputs have values that ranges in $[0,1]$ for the segmentation masks and in $[-1,1]$ for the similarity transformation map since we use $\operatorname{sigmoid}$ and $\operatorname{tanh}$ activation functions, respectively.

The annotation instances are warped using a \emph{Spatial Transformer Network}~\cite{jaderberg2015spatial} that ensures to have differentiable warping operations and allows gradient flow during back-propagation. The warping function performs scale and rotation with respect to the barycenter of the selected annotation instance.
It is noted that the generator $G$ does not receive any information about the separation in instances and about the location of the barycenter of the buildings present in the input mask.
The network, in fact, learns to identify building instances and understands the transformation rules during training.

The regularization network used to refine the segmentation is pre-trained and it is only used during inference.

\section{Experimental Setup}
\label{sec:method}

%(5000 $\times$ 5000 px resolution) 
\subsection{Dataset}
The generator network $G$ and the regularization model are trained in the Inria Aerial Image Labeling Dataset~\cite{maggiori2017dataset} composed of 180 images
($5000 \times 5000$ \SI{}{\px} resolution) of 5 cities from US to Europe.
Two of these images are used as test set.
During training we consider the annotation masks provided in the dataset as ground truth even if some of these images contain misalignments. 

\subsection{Self-supervised training}
The network must receive misaligned and incorrect annotations in order to learn.
Since the dataset is assumed to be made of aligned image pairs some synthetic misalignments and errors must be introduced to alter the ground truth images.
The noise is therefore enhanced by introducing global and instance random translations, rotations and scales.
Random translations have a maximum absolute value of $64$ pixels, while random rotations ranges between \SI{-30}{\degree} and \SI{30}{\degree}.
%$-30^{\circ}$ and $30^{\circ}$.
In order to create the ground truth for the segmentation branches some footprints have also been randomly removed and some others have been injected in the annotation masks.

\section{Results}
\label{sec:result}

\begin{figure}[]
\centering
\includegraphics[width=\linewidth]{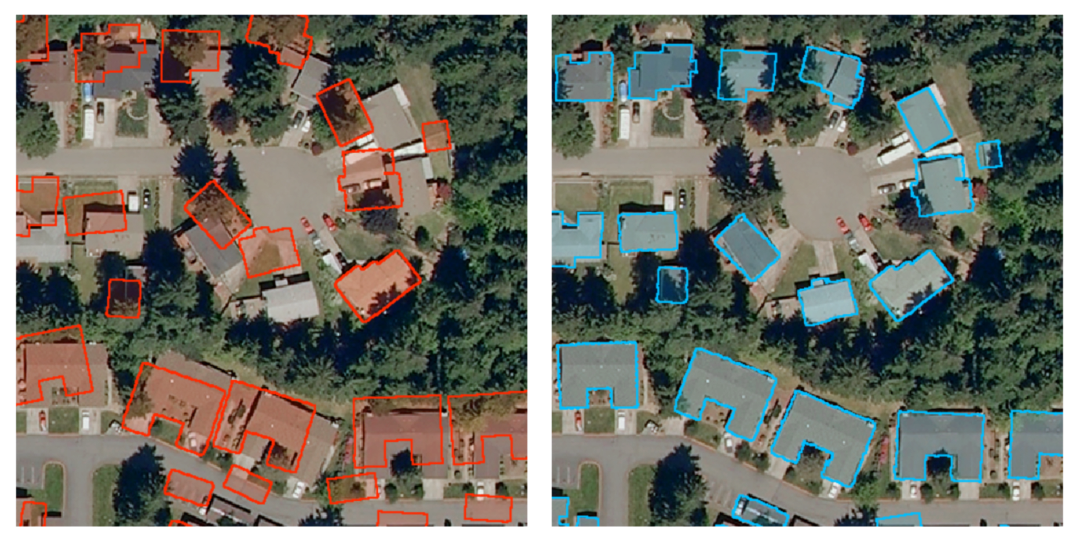}
\caption{Alignment result in kitsap36. Synthetic misaligned annotations on the left. \emph{MapRepair} prediction on the right.}
\label{fig:result_kitsap36}
\end{figure}

\begin{figure}[]
\centering
\includegraphics[width=\linewidth]{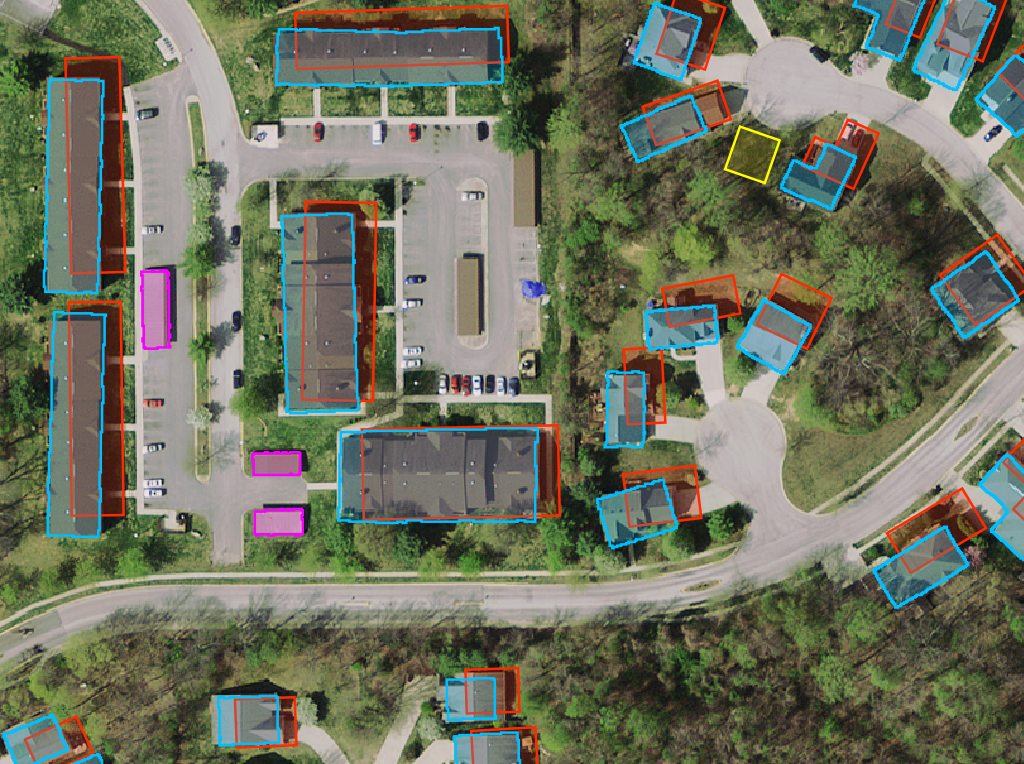}
\caption{Alignment result in bloomington22. Noisy OSM annotations are overlaid in red. \emph{MapRepair} prediction is in cyan. Removed annotations are yellow and segmented buildings are pink.}
\label{fig:result_bloomington22}
\end{figure}

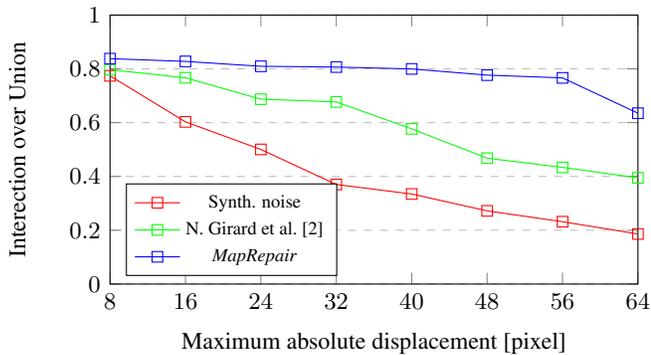
\begin{figure}
\centering
\begin{tikzpicture}
\begin{axis}[
    xlabel={Maximum absolute displacement [pixel]},
    ylabel={Interection over Union},
    xmin=8, xmax=64,
    ymin=0, ymax=1,
    xtick={8,16,24,32,40,48,56,64},
    ytick={0,0.2,0.4,0.6,0.8,1.0},
    legend pos=south west,
    ymajorgrids=true,
    grid style=dashed,
    height=0.6\linewidth,
    width=\linewidth,
    label style={font=\small},
    tick label style={font=\small},
    legend style={font=\scriptsize},
]
\addplot[
    color=red,
    mark=square,
    ]
    coordinates {
    (8,0.774103)(16,0.602877)(24,0.500426)(32,0.37012)(40,0.334734)(48,0.272036)(56,0.231707)(64,0.186039)
    };
    
\addplot[
    color=green,
    mark=square,
    ]
    coordinates {
    (8,0.796829)(16,0.76675)(24,0.687254)(32,0.677082)(40,0.576821)(48,0.467999)(56,0.433674)(64,0.395019)
    };
    
\addplot[
    color=blue,
    mark=square,
    ]
    coordinates {
    (8,0.837913)(16,0.82768)(24,0.809527)(32,0.806633)(40,0.799688)(48,0.776678)(56,0.766416)(64,0.635175)
    };

    \legend{Synth. noise, N. Girard et al.~\cite{girard2019noisy}, \emph{MapRepair}}
\end{axis}
\end{tikzpicture}
\caption{IoU measured with the manually aligned annotations of kitsap36 from the Inria dataset. The plot shows the score of the synthetically misaligned annotations (red) and the score achieved after the correction (blue). The annotations are misaligned with gradually increasing random displacements and with random rotations and scales.}
\label{table:synt}
\end{figure}

\begin{table}[]
\caption{Results in bloomington22 using OSM annotations.}
\begin{tabular}{lc|c|c|c}
\cline{2-5}
\multicolumn{1}{c}{} & \multicolumn{2}{c|}{Alignment} & \multicolumn{2}{c}{Align \& detect} \\ \cline{2-5} 
\multicolumn{1}{c}{} & IoU & Acc & IoU & Acc \\ \hline
\multicolumn{1}{l|}{Misaligned OSM} & 0.5235 & 0.9372 & 0.4739 & 0.9234 \\ \cline{1-1}
\multicolumn{1}{l|}{N. Girard et al.~\cite{girard2019noisy}} & 0.8302 & 0.9813 & 0.7369 & 0.9674 \\ \cline{1-1}
\multicolumn{1}{l|}{\emph{MapRepair} align} & 0.8281 & 0.9812 & 0.7341 & 0.9673 \\ \cline{1-1}
\multicolumn{1}{l|}{\emph{MapRepair} full} & - & - & 0.7914 & 0.9740 \\ \hline
\end{tabular}
\label{table:osm_results}
\end{table}

The method has been evaluated in two Inria images: kitsap36 and bloomington22.
The two images have a resolution of $5000 \times 5000$ pixels and in order to evaluate the full image we split them into $448 \times 448$ patches.
Each patch is individually processed by the network and a $64$ pixels border is discarded due to lack of context information that can lead to the generation of aligned annotations with errors and artifacts.

The kitsap36 image contains 1252 building instances having a wide range of shapes and sizes.
The ground truth provided by the dataset contains several misalignments that are manually corrected in order to evaluate the algorithm prediction.
In this image \emph{MapRepair} correcs the original misaligned ground truth increasing the Intersection over Union (IoU) score from $0.71$ to $0.82$.
Several experiments with synthetic misalignments are conducted in the same test image showing the robustness of the method to heavy annotation displacements.
Building annotations are randomly rotated between $-30\degree$ and $30\degree$ and translated by increasing absolute displacements from $8$ to $64$ pixels.

The results in Figure~\ref{table:synt} show that all the synthetic annotations aligned by \emph{MapRepair} achieve IoU scores around $0.8$.
The best performance in reached with a maximum absolute displacement of $56$ pixels where the network improves the IoU score from $0.23$ to $0.77$ (Figure~\ref{fig:result_kitsap36}).
The efficiency starts dropping with an annotations misalignment of $64$ pixels.

Bloomington22 is an image of the test-set of the Inria dataset, therefore the ground truth is not provided.
For this region OSM provides $771$ building footprints, most of them with severe misalignments.
Furthermore, several construction do not have an OSM annotation.
In order to measure the effectiveness of the correction we manually aligned the footprints and we annotated the unlabelled buildings.
The quantitative and qualitative results in this image are shown in Table~\ref{table:osm_results} and Figure~\ref{fig:result_bloomington22}, respectively.
%In this image \emph{MapRepair} achieves state of the art performance in the alignment task and demonstrates the capability of detecting new buildings exploiting the segmentation and regularization branch.

\section{Conclusions}
We presented \emph{MapRepair}, an approach for cadastre map refinement in satellite images composed of a multi-purpose neural network trained in a self-supervised manner.
The model is capable of generating an aligned cadastre mask predicting a similarity transformation map and warping each object instance independently.
Furthermore, it solves temporal synchronization errors removing unused footprints or segmenting new buildings in the scene.
\emph{MapRepair} achieves comparable or even higher alignment performance with respect to state-of-the-art methods, dealing effectively with heavily distorted annotations.

% References should be produced using the bibtex program from suitable
% BiBTeX files (here: strings, refs, manuals). The IEEEbib.bst bibliography
% style file from IEEE produces unsorted bibliography list.
% -------------------------------------------------------------------------

%\bibliographystyle{IEEEbib}
%\bibliography{strings,refs}
\printbibliography

\end{document}